\documentclass{article}

    \PassOptionsToPackage{numbers, compress}{natbib}
\usepackage[preprint]{neurips_2025}

\usepackage{algorithm}
\usepackage{algpseudocodex} 
\usepackage{bm}

\usepackage[utf8]{inputenc} 
\usepackage[T1]{fontenc}    
\usepackage{hyperref}       
\usepackage{url}            
\usepackage{booktabs}       
\usepackage{amsfonts}       
\usepackage{nicefrac}       
\usepackage{microtype}      
\usepackage{xcolor}         
\usepackage{graphicx}       
\usepackage{amsmath}
\usepackage{subcaption} 
\usepackage{multirow} 

\title{SPAT: Sensitivity-based Multihead-attention Pruning on Time Series Forecasting Models}

\author{%
    Suhan Guo$^{\ast}$, Jiahong Deng\thanks{Equal Contribution}, Mengjun Yi \\
    School of Artificial Intelligence, Nanjing University, China \\
    National Key Laboratory for Novel Software Technology, Nanjing University, China \\
  \texttt{\{shguo, jiahongdeng, mengjunyi\}@smail.nju.edu.cn} \\
  \And
  Furao shen\thanks{Corresponding Author} \\
  School of Artificial Intelligence, Nanjing University, China \\
  National Key Laboratory for Novel Software Technology, Nanjing University, China \\
  \texttt{frshen@nju.edu.cn} \\
  \And
  Jian Zhao \\
  School of Electronic Science and Engineering, Nanjing University, China \\
  National Key Laboratory for Novel Software Technology, Nanjing University, China \\
  \texttt{jianzhao@nju.edu.cn} \\
}

\begin{document}

\maketitle

\begin{abstract}
    Attention-based architectures have achieved superior performance in multivariate time series forecasting but are computationally expensive. Techniques such as patching and adaptive masking have been developed to reduce their sizes and latencies. In this work, we propose a structured pruning method, SPAT (\textbf{S}ensitivity \textbf{P}runer for \textbf{At}tention), which selectively removes redundant attention mechanisms and yields highly effective models. Different from previous approaches, SPAT aims to remove the entire attention module, which reduces the risk of overfitting and enables speed-up without demanding specialized hardware.  We propose a dynamic sensitivity metric, \textbf{S}ensitivity \textbf{E}nhanced \textbf{N}ormalized \textbf{D}ispersion (SEND) that measures the importance of each attention module during the pre-training phase. Experiments on multivariate datasets demonstrate that SPAT-pruned models achieve reductions of 2.842\% in MSE, 1.996\% in MAE, and 35.274\% in FLOPs. Furthermore, SPAT-pruned models outperform existing lightweight, Mamba-based and LLM-based SOTA methods in both standard and zero-shot inference, highlighting the importance of retaining only the most effective attention mechanisms. We have made our code publicly available \url{https://anonymous.4open.science/r/SPAT-6042}.
\end{abstract}


\section{Introduction}
Time series forecasting is the task of predicting future values based on historical observations. It plays a crucial role in various real-world applications, including finance~\cite{liStockMarketForecasting2020}, energy consumption~\cite{sarmasTransferLearningStrategies2022}, traffic monitoring~\cite{wuConnectingDotsMultivariate2020}, healthcare, and epidemic modeling~\cite{dengColaGNNCrosslocationAttention2020}. Accurate forecasting enables informed decision-making, optimizing resource allocation, and mitigating potential risks. Given its widespread impact, improving the accuracy and efficiency of time series forecasting models remains a critical research direction, driving advancements in machine learning and deep learning approaches.

Among the various types of time series forecasting, multivariate time series forecasting focuses on predicting sequences with multiple variables over time~\cite{wuTimesNetTemporal2DVariation2023, nieTimeSeriesWorth2023}. Unlike spatio-temporal~\cite{liuSTAEformerSpatiotemporalAdaptive2023,jiangPDFormerPropagationDelayaware2023} forecasting, multivariate forecasting does not have specific structural constraints, making it a particularly challenging task due to the difficulty of capturing long-range dependencies over extended time horizons. Transformer-based models, leveraging attention mechanisms, have been widely adopted in this domain due to their ability to flexibly model relationships across time steps~\cite{liEnhancingLocalityBreaking2019,zhouInformerEfficientTransformer2021,wuAutoformerDecompositionTransformers2021,liuPyraformerLowcomplexityPyramidal2022,zhouFEDformerFrequencyEnhanced2022}. However, the quadratic complexity of self-attention can significantly increase the computational burden as the lookback window size grows, making efficiency a critical concern~\cite{kitaevReformerEfficientTransformer2020,wangLinformerSelfattentionLinear2020}.

\begin{figure}[t]
     \centering
     \begin{subfigure}[b]{0.35\textwidth}
         \centering
         \includegraphics[width=\textwidth]{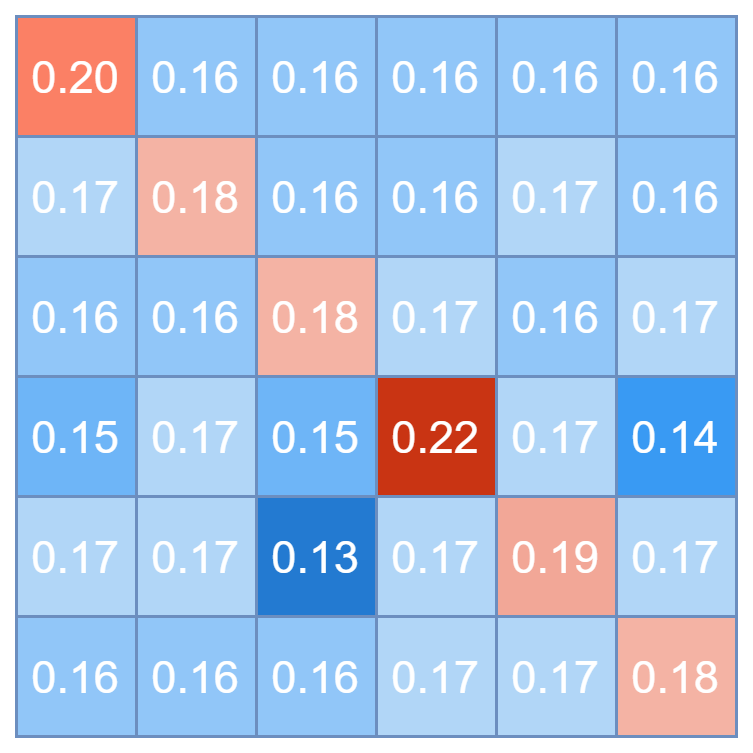}
         \caption{Degenerate attention score}
         \label{fig:deg_att_s}
     \end{subfigure}
     \hspace{15mm}
     \begin{subfigure}[b]{0.35\textwidth}
         \centering
         \includegraphics[width=\textwidth]{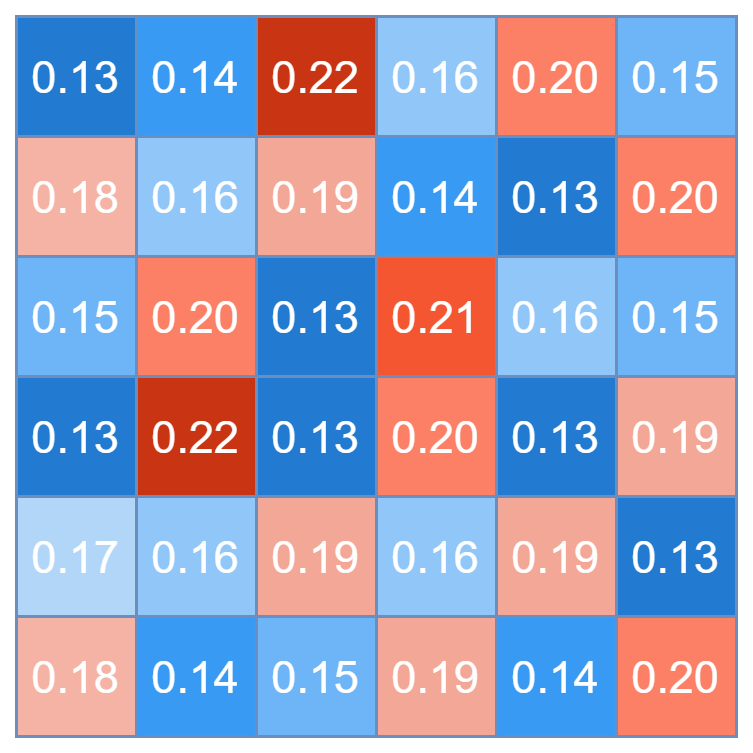}
         \caption{Effective attention score}
         \label{fig:eff_att_s}
     \end{subfigure}
\caption{Attention score visualization: (a) Degenerate vs. (b) Effective. Degenerate attention is equivalent to scaled identity mapping.}
\label{fig:attn_figure}
\end{figure}

Previous work has shown that increasing the lookback window size generally improves forecasting performance by providing the model with a richer temporal context~\cite{nieTimeSeriesWorth2023,jinTimeLLMTimeSeries2024}. However, this improvement comes at the cost of substantial computational complexity. More importantly, not all features at different levels of abstraction in the model require multi-head attention (MHA) mechanisms to form associations~\cite{linMLPCanBe2024,xuIntegratingMambaTransformer2024}. As shown in Figure~\ref{fig:attn_figure}, previous work~\cite{guo2024approximateattentionmlppruning} has found that the degenerated MHA modules are equivalent to scaled identity mapping, rendering some attention computations not only redundant but also detrimental~\cite{luSOFTSEfficientMultivariate2024}, leading to overfitting and reduced generalization ability. 

To address this, we propose a sensitivity-based attention-pruning algorithm that systematically identifies and removes redundant or harmful MHA layers while retaining essential ones. Inspired by previous works~\cite{leeSNIPSingleshotNetwork2019,guoSensitivityPrunerFilterLevel2023}, we define sensitivity as the influence of removing the current attention score matrix on the forecasting performance. By evaluating the dispersion in sensitivity of each attention component, our method prunes less impactful mechanisms, mitigating overfitting and improving generalization.

Furthermore, inspired by recent advances in large language models (LLMs)~\cite{zhouOneFitsAll2023,jinTimeLLMTimeSeries2024}, we explore the zero-shot inference capabilities of attention mechanisms. We evaluate our pruned model on these tasks and find that it retains strong zero-shot prediction abilities, outperforming lightweight competing models that lack sufficient attention mechanisms. These findings highlight the importance of attention layers in capturing complex temporal dependencies, particularly in multivariate time series forecasting tasks that require adaptability with limited supervision.


Our main contributions are summarized as follows:
\begin{itemize}
\item We introduce SPAT, a sensitivity-based attention pruning algorithm, which selectively removes redundant MHA mechanisms, achieving an average reduction of 35.274\% and 28.191\% in FLOPs and parameters. The pruned model outperforms the original by 2.842\% and 1.996\% on average for MSE and MAE.
\item We propose a dispersion-based sensitivity selection strategy, \textbf{S}ensitivity \textbf{E}nhanced \textbf{N}ormalized \textbf{D}ispersion (SEND), that identifies MHA components with minimal performance impact when removed. This selection process is performed adaptively through a pretraining-and-finetuning paradigm.
\item SPAT demonstrates strong zero-shot inference capabilities, surpassing both large language models (LLMs) and light-weight forecasters, highlighting the crucial role of attention mechanisms and the importance of retaining only the most effective ones.
\end{itemize}

\section{Related Work}
\subsection{Transformer-based forecasters}
The Transformer architecture has become central to multivariate time series forecasting due to its ability to capture complex temporal dependencies. To address computational challenges, several models optimize time and space complexity. Informer~\cite{zhouInformerEfficientTransformer2021} improves attention efficiency with probabilistic subsampling, Autoformer~\cite{wuAutoformerDecompositionTransformers2021} uses Auto-Correlation and fast Fourier transforms, and FEDformer~\cite{zhouFEDformerFrequencyEnhanced2022} applies frequency-domain attention with selected components. PatchTST~\cite{nieTimeSeriesWorth2023} introduces a channel-independent patching approach, while Time-LLM~\cite{jinTimeLLMTimeSeries2024} leverages frozen large language models with a trainable adapter for better forecasting efficiency. 

\subsection{Light-weight forecasters} 
To balance performance and efficiency, researchers have explored simpler models such as Convolutional Neural Networks (CNNs)\cite{wangMICNMultiscaleLocal2023, liuSCINetTimeSeries2022, wuTimesNetTemporal2DVariation2023} and Multi-Layer Perceptrons (MLPs)\cite{ekambaramTSMixerLightweightMLPmixer2023,zengAreTransformersEffective2023, luSOFTSEfficientMultivariate2024}. For CNN-based models, TimesNet~\cite{wuTimesNetTemporal2DVariation2023} enhances temporal pattern representation through multi-periodicity, though it struggles with nonstationary datasets. TSLANet~\cite{eldeleTSLANetRethinkingTransformers2024} addresses this by introducing an Adaptive Spectral Block, combining Fourier transforms with local and global filters to improve forecasting accuracy by removing noisy high-frequency components. For MLP-based models, DLinear~\cite{zengAreTransformersEffective2023} achieves superior performance with a single linear layer, outdoing transformer models. The SOFTS model~\cite{luSOFTSEfficientMultivariate2024} builds on this with the STAR module, which aggregates global series representation before redistributing it for encoding local information, boosting efficiency while preserving forecasting accuracy.

\subsection{Attention pruning}
With the rise of Transformer-based models and LLMs, attention pruning has gained attention as a key technique for model compression and acceleration. Pruning algorithms are typically classified into structured and unstructured methods. Structured pruning removes groups of consecutive parameters, improving inference speed on common GPUs, while unstructured pruning removes individual weights, leading to less performance degradation but requiring specialized hardware for acceleration. In NLP, attention pruning has been extensively studied, with methods such as removing attention heads~\cite{michelAreSixteenHeads2019}, pruning FFN hidden dimensions~\cite{liangSuperTicketsPretrained2021}, and eliminating MHA or FFN layers~\cite{xiaStructuredPruningLearns2022}. Additionally, intra-attention pruning~\cite{yangGradientbasedIntraattentionPruning2023} targets structures within attention heads. In computer vision, pruning uninformative attention mechanisms has shown comparable performance to original models~\cite{linMLPCanBe2024}.

However, in time series forecasting, attention pruning remains under-explored. Some studies have investigated adaptive graph sparsification (AGS) to reduce spatial-temporal graph network complexity~\cite{duanLocalisedAdaptiveSpatialtemporal2023}, and PatchTST~\cite{nieTimeSeriesWorth2023} reduces attention complexity by adjusting the lookback window size. Despite these efforts, structured attention pruning is still under-discussed in the context of time series forecasting, which we aim to address in this work.

\section{Methods}

\begin{figure}[t]
     \centering
     \begin{subfigure}[b]{0.35\textwidth}
         \centering
         \includegraphics[width=\textwidth]{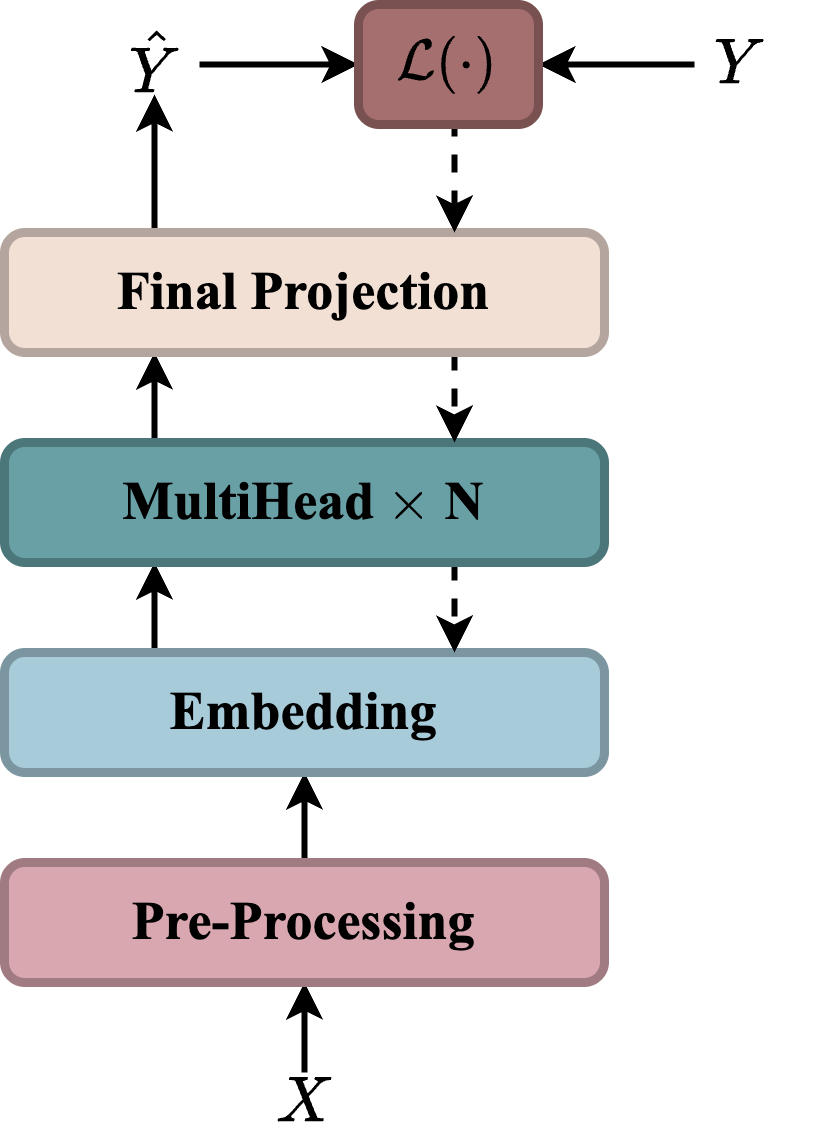}
         \caption{Network Structure}
         \label{fig:network}
     \end{subfigure}
     \hfill
     \begin{subfigure}[b]{0.55\textwidth}
         \centering
         \includegraphics[width=\textwidth]{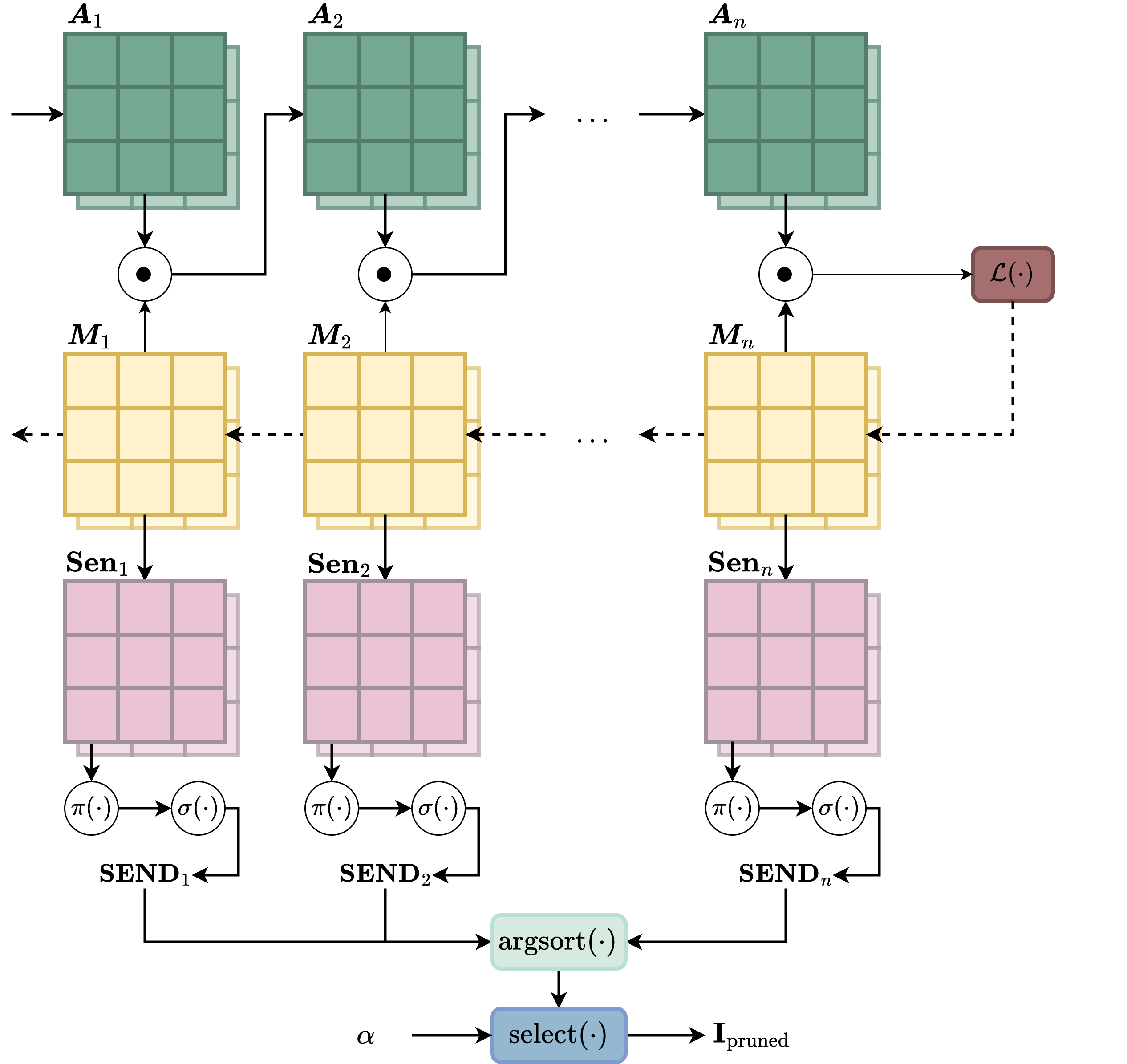}
         \caption{Pruning Algorithm}
         \label{fig:algorithm}
     \end{subfigure}
\caption{Illustration of attention-based models (a) and pruning algorithm workflow (b). Solid arrows indicate feedforward; dashed arrows indicate backpropagation. MultiHead is defined in Eq.\eqref{eq:multihead} and detailed in Appendix Figure\ref{fig:attention}.}
\label{fig:main_figure}
\end{figure}

\subsection{Preliminary}
In multivariate time series forecasting, the goal is to predict future values $\mathrm{\hat{Y}} = \{\mathrm{x}_{L+1}, \dots, \mathrm{x}_{L+T}\} \in \mathbb{R}^{T \times C}$ given past observations $\mathrm{X} = \{\mathrm{x}_1, \dots, \mathrm{x_L}\} \in \mathbb{R}^{L \times C}$, where $L$ is the lookback window and $C$ is the number of channels.

Transformer-based forecasters employ MHA to capture dependencies across time steps and channels. For each attention head $h \in \{1, \dots, H\}$, the attention score matrix $A^{(h)} \in \mathbb{R}^{L \times L}$ is computed using the scaled dot product between the query and key matrices $Q^{(h)}, K^{(h)} \in \mathbb{R}^{L \times d_{\text{head}}}$, followed by a softmax operation $\pi(\cdot)$:
\begin{align} \label{eq:attention_score}
    Q^{(h)} = E_{n-1}^{(h)}W^{(h)}_Q, \quad K^{(h)} = E_{n-1}^{(h)}W^{(h)}_K, \quad
    A^{(h)} = \pi(\frac{Q^{(h)}\left( K^{(h)}\right)^T}{\sqrt{d_{\text{head}}}}).
\end{align}
The $E_{n-1}^{(h)} \in \mathbb{R}^{{L} \times d_{\text{head}}}$ represents the input for the current head, and the weighted sum is obtained using:
\begin{align}
    O^{(h)} = A^{(h)}(E_{n-1}^{(h)}W^{(h)}_V) = A^{(h)}V^{(h)}, \label{eq:attention_value}
\end{align}
where $V^{(h)} \in \mathbb{R}^{L \times d_{\text{head}}}$ is the value. The weighted sums from $H$ heads are concatenated at the head dimension, denoted as $[\cdot]$, and projected with $W_E \in \mathbb{R}^{d_{\text{model}} \times d_{\text{model}}}$, shown as:
\begin{align}
    E_{n} = \left[ O^{(1)}, \dots, O^{(H)} \right] W_E. \label{eq:attention_output}
\end{align}
Summarizing the above equations, we recapitulate the MHA module for Transformer-based models as follows:
\begin{align}
    E_{n} = \text{MultiHead}(E_{n-1}). \label{eq:multihead}
\end{align}

\subsection{SEND}
To develop a method that selectively prunes redundant MHA modules in Transformer-based forecasters, we propose an attention pruning algorithm based on the \textbf{S}ensitivity \textbf{E}nhanced \textbf{N}ormalized \textbf{D}ispersion metric, termed the \textbf{SEND}. The sensitivity of MHA is defined over all $H$ heads. We concatenate the attention score matrices from each head to construct the full attention score matrix $\bm{A}_n$ at layer $n$ in an $N$-layer MHA model:
\begin{align}
    \boldsymbol{A}_n = [A^{(1)},\dots,A^{(H)}] \in \mathbb{R}^{H \times L \times L}. 
\end{align}
We define a connection matrix $\boldsymbol{M}_n \in \{0, 1\}^{H \times L \times L}$, where each entry indicates whether a specific attention score is retained. The pruned attention score matrix is then computed via the Hadamard product: $\boldsymbol{A}’_n = \boldsymbol{A}_n \odot \boldsymbol{M}_n$. Given a dataset $\mathcal{D}$ with $B$ batches, the loss for this attention layer is defined as the average batch loss:
\begin{align}
    \mathcal{L}(\boldsymbol{A}_n \odot \boldsymbol{M}_n ; \mathcal{D}) = \frac{1}{B} \sum_{i=1}^{B} \hat{\mathcal{L}}\left(\bm{A}_n \odot \bm{M}_n ; (\bm{x}_i, \bm{y}_i)\right).  \label{loss_function}
\end{align}
With this matrix, if an attention score in $\bm{A}_n$ at head $H$, attention position $(L_1, L_2)$ is considered redundant, we can simply set the corresponding position in the connection matrix to zero, shown as:
\begin{align}
    \boldsymbol{A}'_{n, (h, l_1, l_2)} &= \boldsymbol{A}_n \odot \boldsymbol{M}_{n, (h, l_1, l_2)}, \\
    \boldsymbol{M}_{n, (h, l_1, l_2)} &=
    \begin{cases}
      0, & h=H, l_1=L_1, l_2=L_2, \nonumber \\
      1, & \text{otherwise}.
    \end{cases}
\end{align}
Inspired by the influence functions from Koh and Liang~\cite{koh2017understanding}, we define the sensitivity of an attention score as the change in loss before and after pruning a given connection. Let $\bm{A}_{n, (H, L_1, L_2)}$ denote the attention score at layer $n$, head $H$, and position $(l_1, l_2)$, with its corresponding binary indicator $m_{n, (h, l_1, l_2)} = 1$ if retained, and $0$ if pruned. To approximate sensitivity as the rate of change in loss $\mathcal{L}$, we relax the binary constraint and define a continuous derivative, denoted as $f_{n, (h, l_1, l_2)}(\boldsymbol{A}_{n}; \mathcal{D})$:
\begin{align}
    &\quad \Delta \mathcal{L}_{n, (h, l_1, l_2)}(\boldsymbol{A}_{n}; \mathcal{D}) \\
    &= \mathcal{L}(\boldsymbol{A}_{n}; \mathcal{D}) - \mathcal{L}(\boldsymbol{A}'_{n, (h, l_1, l_2)}; \mathcal{D}) \nonumber\\ 
    &\approx f_{n, (h, l_1, l_2)}(\boldsymbol{A}_{n}; \mathcal{D}) \nonumber \\
    &= \left. \frac{\partial \mathcal{L}(\boldsymbol{A}_n \odot \boldsymbol{M}_n; \mathcal{D})}{\partial m_{n, (h, l_1, l_2)}} \right\rvert_{\boldsymbol{M}_n=\boldsymbol{1}}  \nonumber \\
    &= \lim_{\delta \to 0} \left. \frac{\mathcal{L}(\boldsymbol{A}_n \odot \boldsymbol{M}_n; \mathcal{D}) - \mathcal{L}(\boldsymbol{A}_n \odot (\boldsymbol{M}_n - \delta \boldsymbol{M}_{n, (h, l_1, l_2)}); \mathcal{D})}{\delta} \right\rvert_{\boldsymbol{M}_n=\boldsymbol{1}}. \nonumber
\end{align}
Using the defined sensitivity measure, we assess the importance of each attention connection by differentiating with respect to the connection mask $\boldsymbol{M}_n$. This yields the sensitivity matrix $\textbf{Sen}_n \in \mathbb{R}^{H \times L \times L}$ for the attention scores $\boldsymbol{A}_n$. It can be efficiently computed in a single forward-backward pass using automatic differentiation in modern frameworks like PyTorch or TensorFlow, as follows:
\begin{equation}
    \label{eq:Sen_generate}
    \bm{\mathrm{Sen}}_n = \frac{\partial \mathcal{L}(\boldsymbol{A}_n \odot \boldsymbol{M}_n; \mathcal{D})}{\partial \boldsymbol{M}_{n}}.
\end{equation}

Based on the chain rule, $\bm{\mathrm{Sen}}_n$ is calculated by applying Hadamard product between a gradient matrix and the attention score matrix. With $\bm{A'}_n = \bm{A}_n \odot \bm{M_n}$, we have:
\begin{align}
    \label{eq:chain_rule}
    \bm{\mathrm{Sen}}_n &= \frac{\partial \mathcal{L}(\bm{A'}_n; \mathcal{D})}{\partial \boldsymbol{M}_{n}} = \frac{\partial \mathcal{L}(\bm{A'}_n; \mathcal{D})}{\partial \bm{A'}_n} \cdot \frac{\partial \bm{A'}_n}{\partial \bm{M}_n} = \frac{\partial \mathcal{L}(\bm{A'}_n; \mathcal{D})}{\partial \bm{A'}_n} \odot \bm{A}_n,
\end{align}
where $\frac{\partial \mathcal{L}(\bm{A'}_n; \mathcal{D})}{\partial \bm{A'}_n} \in \mathbb{R}^{H \times L \times L}$ is the upstream gradient matrix.

Since we focus on the magnitude rather than the direction of gradients, we first apply the absolute value to the sensitivity matrix, removing negative entries. To address potential gradient vanishing or explosion across layers, we normalize the sensitivity values via a row-wise softmax, denoted as $\pi\left( \bm{\mathrm{Sen}}_n\right)$. This transforms raw gradients into a probability distribution, ensuring stability while retaining relative importance across attention positions, as shown below:
\begin{align}
    \label{eq:Sen_norm}
    \pi\left( \bm{\mathrm{Sen}}_n[h, i, j]\right) = \frac{\exp( \left| \bm{\mathrm{Sen}}_n[h,i,j] \right| )}{\sum_{k=1}^L \exp( \left| \bm{\mathrm{Sen}}_n[h,i,k] \right| )}, \quad h \in \{1,...,H\}, \quad i,j \in \{1,...,L\}.
\end{align}

As shown in Equation~\eqref{eq:chain_rule}, the dispersion of the sensitivity matrix is influenced by both the upstream gradient and the attention score matrix. The upstream gradient reflects the impact of the current module on the loss, while the dispersion of the attention scores relates to their effectiveness, as shown in Figure~\ref{fig:attn_figure}. Thus, the influence of an attention layer can be represented by the dispersion of its sensitivity matrix. Hence, we aggregate sensitivities across heads to form a matrix $\overline{\bm{\mathrm{Sen}}}_n \in \mathbb{R}^{L \times L}$:
\begin{align}
    \label{eq:Sen_bar}
    \overline{\bm{\mathrm{Sen}}}_n[i,j] &= \frac{1}{H} \sum_{h=1}^H \pi\left( \bm{\mathrm{Sen}}_n[h,i,j]\right).
\end{align}
We compute the standard deviation of each row $i$ in $\overline{\bm{\mathrm{Sen}}}_n$ to obtain $\sigma_{n, i}$, and then average these values to derive the final score: $\bm{\mathrm{SEND}_n} = \frac{1}{L}\sum_{i=1}^L \sigma_{n, i}$. A higher $\bm{\mathrm{SEND}}_n$ value indicates greater dispersion, suggesting higher effectiveness of the corresponding attention layer. Detailed computation steps are provided in Appendix~\ref{app_sec:implementation}.

\subsection{Ranking and Pruning}
After computing the \textbf{SEND} metric for each of the $N$ attention layers, we rank them in descending order:
\begin{align}
    \bm{I}_{\text{ranked}} = \mathrm{argsort}\left( \{\bm{\mathrm{SEND}}_1, ..., \bm{\mathrm{SEND}}_N\} \right)^{\downarrow},
\end{align}
where $\bm{I}_{\text{ranked}}$ contains the indices of attention layers sorted by their SEND values. Given a pruning ratio $\alpha \in (0, 1)$, we remove the bottom-$K$ layers with the lowest dispersion scores, where:
\begin{align}
    K &= \lceil \alpha N \rceil, \\
    \bm{I}_{\text{pruned}} &= \bm{I}_{\text{ranked}}[N - K + 1 : N].
\end{align}
The complete procedure is summarized in Algorithm~\ref{alg:pipeline}.

\begin{algorithm}[h]
\caption{Training Procedure of SPAT}\label{alg:pipeline}
\begin{algorithmic}
\Require Transformer-based model with $N$ attention layers, state set $S$, candidate set $C$, \\
number of layers to remove $K$, sparse mask $M$, loss $\mathcal{L}$.
\Ensure Simplified model with attention layers $S$ removed.

\State $S \gets \emptyset,\ C \gets \{\mathrm{Attn}_1, \mathrm{Attn}_2, \dots, \mathrm{Attn}_N\},\ M \gets \mathbf{1}$

\For {each layer $n$}
    \State $\bm{\mathrm{Sen}}_n \gets \mathrm{Backprop} (\mathcal{L}(\bm{\mathrm{A}}_n \odot \bm{\mathrm{M}}_n))$ \Comment{Refer to Eq. \ref{eq:Sen_generate}} 
    \State $\bm{\mathrm{Sen}}_n^{\mathrm{(norm)}}[h,i,j] \gets \pi\left( \bm{\mathrm{Sen}}_n[h,i,j]\right), \forall h,i,j$\Comment{Refer to Eq. \ref{eq:Sen_norm}} 
    \State $\overline{\bm{\mathrm{Sen}}}_n = \mathbb{E}_h\left[\bm{\mathrm{Sen}}_n^{\mathrm{(norm)}}[h,:,:]\right]$ \Comment{Refer to Eq. \ref{eq:Sen_bar}} 
    \State $\bm{\mathrm{SEND}}_n = \mathbb{E}_{i}  \left[ \sigma_{n,i}\left(\overline{\bm{\mathrm{Sen}}}_{n}\right) \right]$\Comment{average standard deviation of each row}
\EndFor
\textbf{end for}\\

\State  $\bm{I}_{\text{ranked}} \gets \mathrm{argsort} \left( \{\bm{\mathrm{SEND}}_1, ..., \bm{\mathrm{SEND}}_N\} \right)$ \Comment{Find optimal layers to remove}
\State  $\bm{I}_{\text{pruned}} \gets \bm{I}_{\text{ranked}}[N - K + 1 : N]$ 
\State $ S \gets S \cup\{\mathrm{Attn}_{\bm{I}_{\text{pruned}}}\},\quad C \gets C \setminus \{\mathrm{Attn}_{\bm{I}_{\text{pruned}}}\}$\Comment{Update sets}
\State \textbf{return} Simplified model with attention layers $S$ removed
\end{algorithmic}
\end{algorithm}

\section{Experiments}
We empirically evaluate our SPAT on multivariate time series forecasting tasks. The experiments show that many MHA modules are redundant, and removing them improves forecasting accuracy compared to SOTA baselines. Moreover, the retained attention modules are essential for preserving zero-shot inference capabilities, underscoring their importance for generalization.

\subsection{Experimental set-up} \label{sec:set_up}
We evaluate our proposed SPAT on 8 real-world datasets: ETT (4 subsets), Traffic, Electricity, Weather, and ILI~\cite{zhouInformerEfficientTransformer2021, wuAutoformerDecompositionTransformers2021}. Further details on the datasets can be found in Appendix~\ref{app_sec:dset}.

We compare our pruned model against recent lightweight methods such as DLinear~\cite{zengAreTransformersEffective2023}, TSLANet~\cite{eldeleTSLANetRethinkingTransformers2024}, and SOFTS~\cite{luSOFTSEfficientMultivariate2024}, as well as LLM-based methods and MAMBA-based models like Time-LLM~\cite{jinTimeLLMTimeSeries2024} and Bi-Mamba+~\cite{liangBiMambaBidirectionalMamba2024}.

For benchmarking, we use the setup from PatchTST~\cite{nieTimeSeriesWorth2023}, with a 336-length lookback window and varying prediction horizons $(T)$: {24, 36, 48, 60} for ILI and {96, 192, 336, 720} for the others. Performance is measured using Mean Squared Error (MSE) and Mean Absolute Error (MAE).

We use the ADAM optimizer with an initial learning rate of $3 \times 10^{-4}$. We follow the original implementation in PatchTST and iTransformer. Implementation details are in Appendix~\ref{app_sec:implementation}.

\begin{table}[t]
\centering
\caption{Performance of PatchTST (Top) and iTransformer (Bottom) on multivariate datasets. Results are averaged over 4 forecasting horizons. Columns represent (1) the original model, (2) the pruned model, and (3) the improvement between them. MSE and MAE improvements are highlighted in bold.}
\label{tab:PatchTST_itrans_main}
\resizebox{\textwidth}{!}{\begin{tabular}{l|rr|rr|rrrr}
\toprule
\multicolumn{1}{l|}{} & \multicolumn{2}{c|}{Original Model} & \multicolumn{2}{c|}{Pruned Model} & \multicolumn{4}{c}{Improvement} \\
\midrule
\multicolumn{1}{l|}{} & \multicolumn{1}{c}{MSE} & \multicolumn{1}{c|}{MAE} & \multicolumn{1}{c}{MSE} & \multicolumn{1}{c|}{MAE} & \multicolumn{1}{c}{MSE($\uparrow$)} & \multicolumn{1}{c}{MAE($\uparrow$)} & \multicolumn{1}{c}{FLOPs($\downarrow$)} & \multicolumn{1}{c}{Params($\downarrow$)} \\
\midrule
Weather & $0.229\pm0.000$ & $0.264\pm0.000$ & $0.229\pm0.000$ & $0.264\pm0.000$ & 0.000\% & 0.000\% & 16.210\% & 2.984\%  \\
Traffic & $0.389\pm0.001$ & $0.262\pm0.001$ & $0.389\pm0.001$ & $0.260\pm0.001$ & 0.000\% & \textbf{-0.763\%} & 16.210\% & 2.984\% \\
Electricity & $0.163\pm0.000$ & $0.255\pm0.000$ & $0.162\pm0.000$ & $0.254\pm0.000$ & \textbf{-0.613\%} & \textbf{-0.392\%} & 16.209\% & 2.984\%  \\
ILI & $1.728\pm0.024$ & $0.885\pm0.005$ & $1.587\pm0.048$ & $0.850\pm0.006$ & \textbf{-8.160\%} & \textbf{-3.955\%} & 11.685\% & 2.390\% \\
ETTh1 & $0.415\pm0.003$ & $0.43\pm0.003$ & $0.413\pm0.003$ & $0.427\pm0.002$ & \textbf{-0.482\%} & \textbf{-0.698\%} & 9.593\% & 0.447\%  \\
ETTh2 & $0.331\pm0.001$ & $0.381\pm0.001 $& $0.330\pm0.000$ & $0.380\pm0.000$ & \textbf{-0.302\%} & \textbf{-0.262\%} & 9.593\% & 0.447\%  \\
ETTm1 & $0.352\pm0.001$ & $0.382\pm0.000$ & $0.350\pm0.001$ & $0.380\pm0.000$ & \textbf{-0.568\%} & \textbf{-0.524\%} & 16.210\% & 2.984\% \\
ETTm2 & $0.256\pm0.000$ & $0.314\pm0.000$ & $0.256\pm0.000$ & $0.314\pm0.000$ & 0.000\% & 0.000\% & 16.210\% & 2.984\% \\
\midrule
Weather & $0.238\pm0.000$ & $0.273\pm0.000$ & $0.228\pm0.000$ & $0.265\pm0.001$ & \textbf{-4.202\%} & \textbf{-2.930\%} & 62.574\% & 62.047\% \\
Traffic & $0.386\pm0.001$ & $0.273\pm0.000$ & $0.377\pm0.001$ & $0.268\pm0.000$ & \textbf{-2.332\%} & \textbf{-1.832\%} & 56.942\% & 47.342\% \\
Electricity & $0.165\pm0.001$ & $0.259\pm0.002$ & $0.161\pm0.000$ & $0.252\pm0.001$ & \textbf{-2.424\%} & \textbf{-2.703\%} & 68.254\% & 62.047\% \\
ILI & $2.202\pm0.014$ & $1.021\pm0.002$ & $2.111\pm0.024$ & $0.989\pm0.004$ & \textbf{-4.133\%} & \textbf{-3.134\%} & 64.883\% & 64.616\% \\
ETTh1 & $0.471\pm0.002$ & $0.466\pm0.001$ & $0.426\pm0.002$ & $0.438\pm0.001$ & \textbf{-9.554\%} & \textbf{-6.009\%} & 58.456\% & 58.193\% \\
ETTh2 & $0.384\pm0.004$ & $0.414\pm0.002$ & $0.361\pm0.001$ & $0.399\pm0.000$ & \textbf{-5.990\%} & \textbf{-3.623\%} & 47.117\% & 46.203\% \\
ETTm1 & $0.368\pm0.001$ & $0.395\pm0.000$ & $0.354\pm0.001$ & $0.382\pm0.000$ & \textbf{-3.804\%} & \textbf{-3.291\%} & 47.117\% & 46.203\% \\
ETTm2 & $0.275\pm0.001$ & $0.331\pm0.001$ & $0.267\pm0.001$ & $0.325\pm0.000$ & \textbf{-2.909\%} & \textbf{-1.813\%} & 47.117\% & 46.203\% \\
\bottomrule
\end{tabular}}
\end{table}

\subsection{Results}
We present our key experiments in three sections: ``Original vs. Pruned Models,'' comparing the pruned models to their original versions; ``Pruned vs. Benchmark Models,'' comparing the pruned models with benchmarks; and ``Zero-shot Inference,'' focusing on zero-shot performance. To validate our approach, we first train all models on their respective datasets to establish baseline performance. Then, we apply SPAT to prune attention modules and fine-tune the pruned models. All models—original, pruned, and benchmark—are evaluated under identical conditions. We report the improvement in MSE and MAE for the pruned models, averaged over five runs. Additionally, we track FLOPs ($\downarrow$) and Params ($\downarrow$) to show the reductions in inference complexity and model size, respectively.

\paragraph{Original vs. Pruned Models} 
As shown in the top row of Table~\ref{tab:PatchTST_itrans_main}, the pruned PatchTST achieves a 1.266\% reduction in MSE and 0.824\% in MAE, along with 13.990\% and 2.275\% reductions in FLOPs and parameters, respectively. Performance remains equal or better across all datasets, with notable gains on ILI, highlighting that illness-related data may be more prone to overfitting due to redundant attention. The consistent drop in FLOPs and parameters is attributed to the fixed lookback window and PatchTST’s channel-independent design.

The bottom row in Table~\ref{tab:PatchTST_itrans_main} shows similar trends for iTransformer, which achieves 4.418\% and 3.167\% reductions in MSE and MAE, and over 50\% drops in FLOPs and parameters. The pruned iTransformer consistently improves performance across all datasets, indicating that redundant attention harms generalization regardless of the number of variables in the dataset. While iTransformer benefits more from pruning in terms of efficiency, pruned PatchTST still delivers stronger forecasting accuracy overall in the current experimental setup.

\paragraph{Pruned vs. Benchmark Models}

\begin{table}[t]
\centering
\caption{Performance comparison with baseline models on multivariate datasets. Results are averaged over 4 forecasting horizons. Best MSE and MAE are highlighted in \textbf{bold}, the second best in \underline{underline}.}
\label{tab:comparison_main}
\resizebox{\textwidth}{!}{\begin{tabular}{lcccccccccccccc}
\toprule
 \multicolumn{1}{c}{Type} & \multicolumn{6}{c}{Light-weight} & \multicolumn{2}{c}{LLM-based} & \multicolumn{2}{c}{Mamba-based} & \multicolumn{4}{c}{Ours} \\ \cmidrule(lr){1-1} \cmidrule(lr){2-7} \cmidrule(lr){8-9} \cmidrule(lr){10-11} \cmidrule(lr){12-15}
 \multicolumn{1}{c}{Model} & \multicolumn{2}{c}{DLinear} & \multicolumn{2}{c}{TSLANet} & \multicolumn{2}{c}{SOFTS} & \multicolumn{2}{c}{Time-LLM} & \multicolumn{2}{c}{Bi-Mamba+} & \multicolumn{2}{c}{Pruned PatchTST} & \multicolumn{2}{c}{Pruned iTransformer} \\ 
 \cmidrule(lr){1-1} \cmidrule(lr){2-3} \cmidrule(lr){4-5} \cmidrule(lr){6-7} \cmidrule(lr){8-9} \cmidrule(lr){10-11} \cmidrule(lr){12-13} \cmidrule(lr){14-15}
 \multicolumn{1}{c}{Metric} & MSE & \multicolumn{1}{c|}{MAE} & MSE & \multicolumn{1}{c|}{MAE} & MSE & \multicolumn{1}{c|}{MAE} & MSE & \multicolumn{1}{c|}{MAE} & MSE & \multicolumn{1}{c|}{MAE} & MSE & \multicolumn{1}{c|}{MAE} & MSE & MAE \\ \midrule
Weather & 0.246 & 0.299 & 0.246 & 0.279 & 0.246 & 0.278 & 0.238 & 0.274 & 0.238 & 0.275 & \underline{0.229} & \textbf{0.264} & \textbf{0.228} & \underline{0.265} \\
Traffic & 0.454 & 0.328 & 0.406 & 0.274 & \underline{0.388} & \underline{0.266} & 0.410 & 0.291 & 0.389 & 0.281 & 0.389 & \textbf{0.260} & \textbf{0.377} & 0.268 \\
Electricity & 0.166 & 0.264 & 0.162 & 0.255 & 0.163 & 0.257 & 0.172 & 0.271 & 0.167 & 0.266 & \underline{0.162} & \underline{0.254} & \textbf{0.161} & \textbf{0.252} \\
ILI & 2.201 & 1.054 & 1.859 & 0.908 & 2.065 & 0.976 & 2.198 & 0.972 & \underline{1.796} & \underline{0.891} & \textbf{1.587} & \textbf{0.850} & 2.111 & 0.989 \\
ETTh1 & 0.445 & 0.458 & \underline{0.422} & \underline{0.433} & 0.426 & 0.437 & 0.442 & 0.448 & 0.436 & 0.437 & \textbf{0.413} & \textbf{0.427} & 0.426 & 0.438 \\
ETTh2 & 0.482 & 0.472 & \underline{0.331} & \underline{0.384} & 0.380 & 0.411 & 0.374 & 0.409 & 0.378 & 0.411 & \textbf{0.330} & \textbf{0.380} & 0.361 & 0.399 \\
ETTm1 & 0.360 & 0.382 & \textbf{0.343} & \textbf{0.379} & 0.364 & 0.391 & 0.359 & 0.388 & 0.362 & 0.389 & \underline{0.350} & \underline{0.380} & 0.354 & 0.382 \\
ETTm2 & 0.281 & 0.344 & 0.277 & 0.333 & 0.282 & 0.331 & 0.268 & \underline{0.324} & 0.270 & 0.326 & \textbf{0.256} & \textbf{0.314} & \underline{0.267} & 0.325 \\\bottomrule
\end{tabular}}
\end{table}

As shown in Table~\ref{tab:comparison_main}, our pruned model outperforms SOTA lightweight, LLM-based, and Mamba-based multivariate time series forecasting models across all datasets, with the exception of ETTm1. On high-dimensional datasets such as Weather, Traffic, and Electricity (with $>10$ variables), both pruned PatchTST and iTransformer deliver leading performance. For lower-dimensional datasets, pruned PatchTST remains more robust. TSLANet is a strong baseline, ranking second on most datasets and outperforming others on ETTm1, likely due to the increased overfitting risk of attention-based models on lower-dimensional datasets like ETT. In contrast, Time-LLM and Mamba-based models do not seem to offer robust performance. 

\paragraph{Zero-shot Inference} 
To assess the role of essential attention modules, we evaluate zero-shot performance by testing models on unseen datasets $\star$ after training on distinct datasets $\dagger$, following the transfer setting of Time-LLM~\cite{jinTimeLLMTimeSeries2024}. The results in Table~\ref{tab:comparison_zeroshot} show that the pruned PatchTST retains its zero-shot capability, while the pruned iTransformer performs adequately. Notably, the zero-shot performance of the pruned PatchTST exceeds that of the second-best model by an average of 5.253\% in MSE and 3.005\% in MAE. Among benchmarks, Time-LLM achieves the most second-best results, further emphasizing the importance of retaining key MHA modules for cross-dataset generalization.

\begin{table}[t]
\centering
\caption{Zero-shot performance comparison on multivariate datasets. Left of $\rightarrow$: training dataset; right: inference dataset. Results are averaged over 4 forecasting horizons. Best MSE and MAE are highlighted in \textbf{bold}, the second best in \underline{underline}.}
\label{tab:comparison_zeroshot}
\resizebox{\textwidth}{!}{\begin{tabular}{lcccccccccccccc}
\toprule
 \multicolumn{1}{c}{Type} & \multicolumn{6}{c}{Light-weight} & \multicolumn{2}{c}{LLM-based} & \multicolumn{2}{c}{Mamba-based} & \multicolumn{4}{c}{Ours} \\ \cmidrule(lr){1-1} \cmidrule(lr){2-7} \cmidrule(lr){8-9} \cmidrule(lr){10-11} \cmidrule(lr){12-15}
 \multicolumn{1}{c}{Model} & \multicolumn{2}{c}{DLinear} & \multicolumn{2}{c}{TSLANet} & \multicolumn{2}{c}{SOFTS} & \multicolumn{2}{c}{Time-LLM} & \multicolumn{2}{c}{Bi-Mamba+} & \multicolumn{2}{c}{Pruned PatchTST} & \multicolumn{2}{c}{Pruned iTransformer} \\ 
 \cmidrule(lr){1-1} \cmidrule(lr){2-3} \cmidrule(lr){4-5} \cmidrule(lr){6-7} \cmidrule(lr){8-9} \cmidrule(lr){10-11} \cmidrule(lr){12-13} \cmidrule(lr){14-15}
 \multicolumn{1}{c}{Metric} & MSE & \multicolumn{1}{c|}{MAE} & MSE & \multicolumn{1}{c|}{MAE} & MSE & \multicolumn{1}{c|}{MAE} & MSE & \multicolumn{1}{c|}{MAE} & MSE & \multicolumn{1}{c|}{MAE} & MSE & \multicolumn{1}{c|}{MAE} & MSE & MAE \\ \midrule
$ETTh1 \rightarrow ETTh2$ & 0.538 & 0.505 & 0.359 & \underline{0.390} & 0.360 & 0.399 & 0.360 & 0.399 & \underline{0.353} & 0.392 & \textbf{0.334} & \textbf{0.381} & 0.363 & 0.395 \\
$ETTh1 \rightarrow ETTm2$ & 0.476 & 0.485 & 0.320 & 0.359 & 0.330 & 0.368 & 0.308 & 0.360 & \underline{0.305} & \underline{0.355} & \textbf{0.295} & \textbf{0.349} & 0.308 & 0.362 \\
$ETTh2 \rightarrow ETTh1$ & 0.569 & 0.518 & \underline{0.525} & \underline{0.493} & 0.629 & 0.558 & 0.584 & 0.528 & 0.651 & 0.533 & \textbf{0.496} & \textbf{0.482} & 0.596 & 0.542 \\
$ETTh2 \rightarrow ETTm2$ & 0.428 & 0.446 & 0.329 & 0.369 & 0.321 & 0.372 & \underline{0.305} & \underline{0.360} & 0.308 & 0.362 & \textbf{0.285} & \textbf{0.342} & 0.312 & 0.366 \\
$ETTm1 \rightarrow ETTh2$ & 0.471 & 0.476 & 0.427 & 0.435 & 0.406 & 0.427 & \underline{0.379} & \underline{0.414} & 0.392 & 0.423 & \textbf{0.360} & \textbf{0.402} & 0.391 & 0.419 \\
$ETTm1 \rightarrow ETTm2$ & 0.290 & 0.352 & 0.301 & 0.338 & 0.282 & 0.331 & 0.274 & 0.326 & 0.273 & 0.327 & \textbf{0.262} & \textbf{0.317} & \underline{0.268} & \underline{0.323} \\
$ETTm2 \rightarrow ETTh2$ & 0.386 & 0.420 & 0.390 & 0.418 & 0.441 & 0.442 & 0.379 & \underline{0.412} & \underline{0.378} & \underline{0.412} & \textbf{0.356} & \textbf{0.396} & 0.450 & 0.458 \\
$ETTm2 \rightarrow ETTm1$ & 0.508 & 0.463 & \underline{0.476} & \underline{0.448} & 0.535 & 0.489 & 0.485 & 0.456 & 0.491 & 0.454 & \textbf{0.444} & \textbf{0.430} & 0.563 & 0.503\\ \bottomrule
\end{tabular}}
\end{table}

\subsection{Model efficiency} \label{sec:model_efficiency}
\begin{figure}[t]
     \begin{subfigure}[t]{0.3\textwidth}
         \centering
         \includegraphics[width=1.35\textwidth]{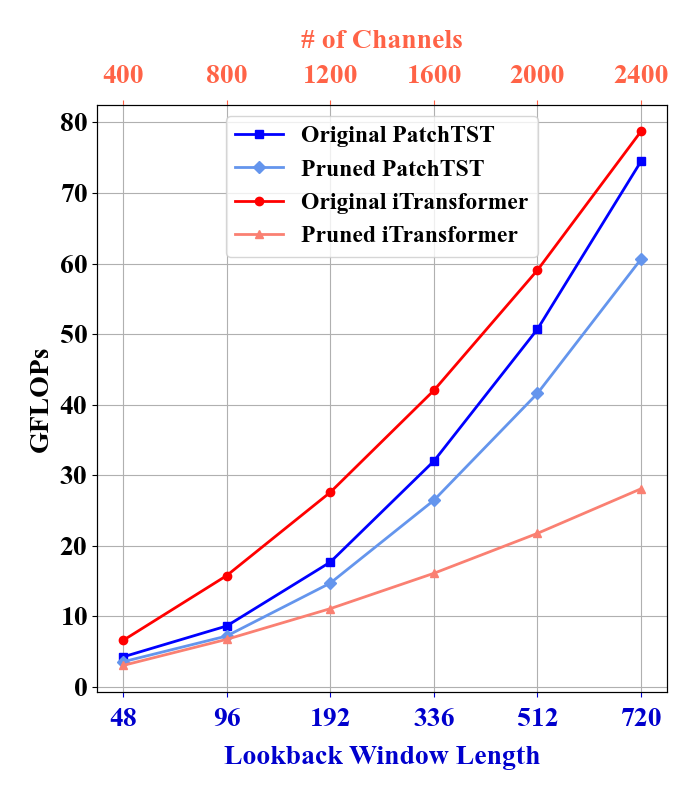}
         \caption{}
         \label{fig:lookback_gflops}
     \end{subfigure}
     \hspace{16mm}
     \begin{subfigure}[t]{0.5\textwidth}
         \centering
         \includegraphics[width=1.1\textwidth]{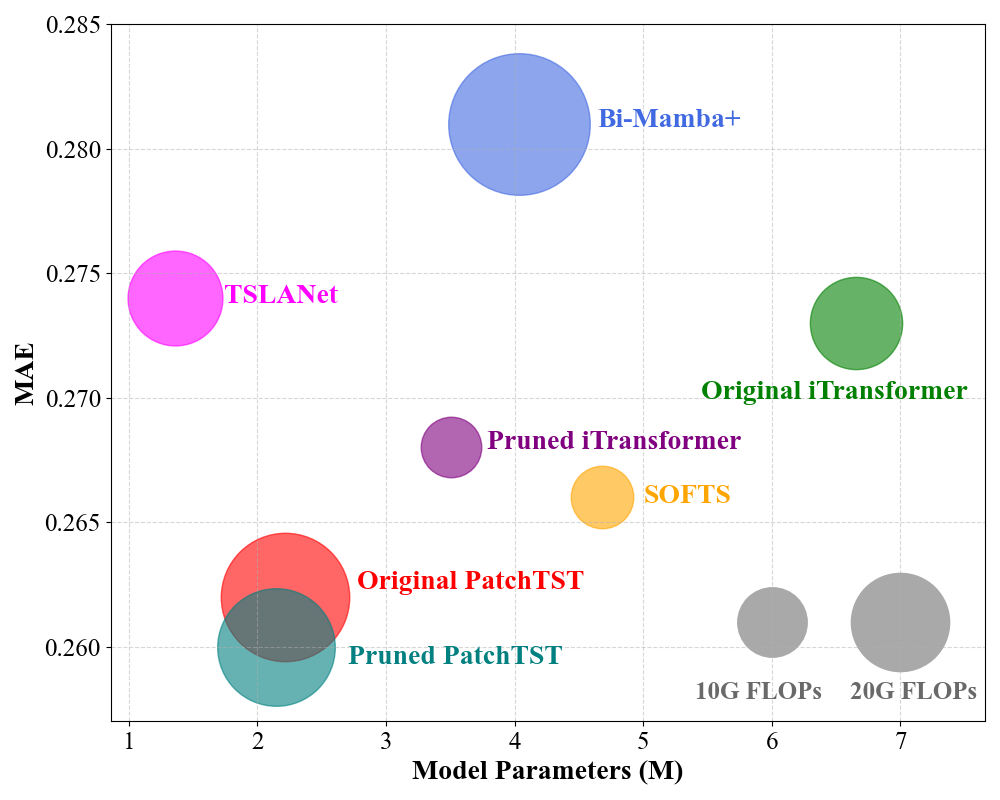}
         \caption{}
         \label{fig:model_params}
     \end{subfigure}
\caption{Model efficiency comparison with average results over 4 forecasting horizons on Traffic. The batch size is set to 1. Figure~\ref{fig:lookback_gflops} shows that PatchTST FLOPs vary with lookback length (bottom x-axis); iTransformer FLOPs vary with channel count (top x-axis).}
\label{fig:model_efficiency}
\end{figure}

Due to the channel-independent design of PatchTST, its MHA complexity is primarily dependent on the lookback window length $L$. In contrast, the MHA in iTransformer is solely dependent on the number of channels. As illustrated in Figure~\ref{fig:lookback_gflops}, the pruned PatchTST exhibits reduced sensitivity to increases in $L$, while the pruned iTransformer shows greater robustness to increases in the number of channels. Figure~\ref{fig:model_params} plots model size against MAE. Both pruned PatchTST and iTransformer achieve lower error with fewer parameters. Compared to light-weight models like TSLANet and SOFTS, pruned PatchTST offers better accuracy with a similar model size. We exclude Time-LLM and Dlinear due to their large model size and inadequate performance, respectively, which would obscure differences among the other models.

\subsection{Ablation Studies}

\begin{table}[t]
\centering
\caption{Effect of pruning ratios on PatchTST (Top) and iTransformer (Bottom). Higher $\alpha$ indicates more pruned attention modules. ‘0.3 vs. 0.9’ shows the percentage change in metrics between $\alpha=0.3$ and $\alpha=0.9$.}
\label{tab:abl_pruning_ratio}
\resizebox{\textwidth}{!}{\begin{tabular}{@{}lllllllllllll@{}}
\toprule
  \multicolumn{1}{c}{\multirow{2}{*}{Ratio}} & \multicolumn{2}{c}{Weather} & \multicolumn{2}{c}{Traffic} & \multicolumn{2}{c}{Electricity} & \multicolumn{2}{c}{ILI} & \multicolumn{2}{c}{ETTh1} & \multicolumn{2}{c}{ETTm1} \\ 
 \cmidrule(lr){2-3} \cmidrule(lr){4-5} \cmidrule(lr){6-7}  \cmidrule(lr){8-9} \cmidrule(lr){10-11} \cmidrule(lr){12-13}
 \multicolumn{1}{c}{} & \multicolumn{1}{c}{MSE} & \multicolumn{1}{c}{MAE} & \multicolumn{1}{c}{MSE} & \multicolumn{1}{c}{MAE} & \multicolumn{1}{c}{MSE} & \multicolumn{1}{c}{MAE} & \multicolumn{1}{c}{MSE} & \multicolumn{1}{c}{MAE} & \multicolumn{1}{c}{MSE} & \multicolumn{1}{c}{MAE} & \multicolumn{1}{c}{MSE} & \multicolumn{1}{c}{MAE} \\ \midrule
 $\alpha=0$ & \textbf{0.229} & \textbf{0.264} & \textbf{0.389} & 0.262 & 0.163 & 0.255 & 1.728 & 0.885 & 0.415 & 0.430 & 0.352 & 0.382 \\
 $\alpha=0.3$ & \textbf{0.229} & \textbf{0.264} & \textbf{0.389} & \textbf{0.260} & \textbf{0.162} & \textbf{0.254} & \textbf{1.587} & \textbf{0.850} & \textbf{0.413} & \textbf{0.427} & \textbf{0.350} & \textbf{0.380} \\
 $\alpha=0.9$ & 0.233 & 0.266 & 0.411 & 0.274 & 0.164 & 0.257 & 1.616 & 0.856 & 0.420 & 0.432 & 0.353 & 0.386 \\
 0.3 vs. 0.9 & 1.747\% & 0.758\% & 5.656\% & 5.385\% & 1.235\% & 1.181\% & 1.827\% & 0.706\% & 1.695\% & 1.171\% & 0.857\% & 1.579\% \\ \midrule
 $\alpha=0$ & 0.238 & 0.273 & 0.386 & 0.273 & 0.165 & 0.259 & 2.202 & 1.021 & 0.471 & 0.466 & 0.368 & 0.395 \\
 $\alpha=0.3$ & 0.237 & 0.272 & 0.375 & 0.266 & 0.165 & 0.259 & 2.256 & 1.029 & 0.454 & 0.455 & 0.363 & 0.391 \\
 $\alpha=0.9$ & \textbf{0.228} & \textbf{0.265} & \textbf{0.377} & \textbf{0.268} & \textbf{0.161} & \textbf{0.252} & \textbf{2.111} & \textbf{0.989} & \textbf{0.426} & \textbf{0.438} & \textbf{0.354} & \textbf{0.382} \\
 0.3 vs. 0.9 & -3.797\% & -2.574\% & 0.533\% & 0.752\% & -2.424\% & -2.703\% & -6.427\% & -3.887\% & -6.167\% & -3.736\% & -2.479\% & -2.302\% \\ \bottomrule
\end{tabular}}
\end{table}

We use the same prediction horizon setting as in main experiments. All the results are averaged over four horizons. Unless otherwise noted, the lookback window length is set to 336 by default.

\paragraph{Comparison of different pruning ratios} 
Table~\ref{tab:abl_pruning_ratio} confirms that varying the pruning ratio $\alpha$ affects performance, but the overall trend remains consistent. For PatchTST, $\alpha=0.9$, which removes all MHA modules, leads to a performance drop with an increase of 0.540\% in MSE and 0.723\% in MAE, while $\alpha=0.3$, which removes one MHA, yields optimal results. In contrast, iTransformer achieves the best performance when three MHA modules are pruned, and even with $\alpha=0.3$, it improves over the original model by 1.165\% in MSE and 0.924\% in MAE, suggesting higher redundancy in its MHA layers under current settings.

\begin{figure}[h!]
    \centering
    \includegraphics[width=\textwidth]{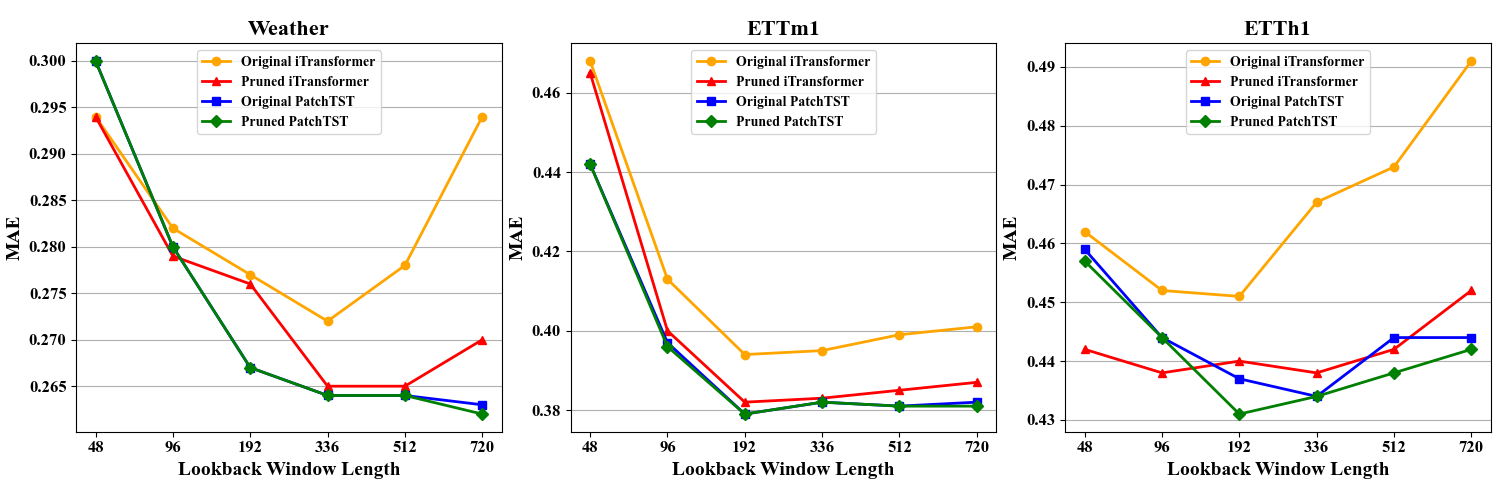}
    \caption{Influence of lookback window lengths $L$}
    \label{fig:lookback_sizes}
\end{figure}

\paragraph{Influence of lookback window length}
We assess the impact of lookback window length on both pruned and original models in Figure~\ref{fig:lookback_sizes}. The results show a consistent trend: increasing the lookback window length generally improves forecasting performance. However, when the window length reaches or exceeds 336, both the pruned and original iTransformer exhibit degraded performance, particularly on the ETTh1 dataset. Notably, the pruned iTransformer displays reduced performance fluctuation, aligning with our earlier observation that the MHA modules in its architecture are somewhat redundant.

\section{Conclusion}
While multi-head attention mechanisms introduce computational costs that scale quadratically with the lookback window length and/or number of channels, they also provide crucial information for multivariate time series forecasting, especially in zero-shot inference scenarios. Existing pruning approaches largely focus on unstructured attention score pruning, which often requires specialized hardware to realize inference acceleration. In this work, we propose SPAT, a sensitivity-based pruning algorithm that leverages the SEND importance metric to identify and remove redundant attention modules in Transformer-based forecasters. Our empirical results show that SPAT can significantly reduce both model size and inference cost while preserving or even improving forecasting performance. In comparison with lightweight, LLM-based, and Mamba-based baselines, the pruned models, particularly PatchTST, achieve superior results on most datasets. Zero-shot transfer evaluations further highlight the importance of retaining key attention mechanisms for model generalization. Overall, this study demonstrates that selective pruning of attention modules offers an effective path toward building efficient and high-performing time series forecasting models.

\section*{Acknowledgments}
This work was supported in part by the STI 2030-Major Projects of China under Grant 2021ZD0201300, the National Science Foundation of China under Grant 62276127, and the Fundamental Research Funds for the Central Universities under Grant 2024300394.

\bibliography{attention_prune}

\appendix
\section{Datasets Description} \label{app_sec:dset}
The detailed information for the dataset is presented below:
\begin{enumerate}
    \item The Weather dataset~\cite{wuTimesNetTemporal2DVariation2023} includes one-year records from 21 meteorological stations located in Germany, with a sampling rate of 10 minutes.
    \item The traffic dataset~\cite{wuTimesNetTemporal2DVariation2023} describes the road occupancy rates. It contains the hourly data recorded by the 762 sensors of San Francisco freeways from 2015 to 2016.
    \item The Electricity dataset~\cite{wuTimesNetTemporal2DVariation2023} comprises two-year records of electricity consumption from 321 customers, measured at a 1-hour sampling rate.
    \item The influenza-like illness (ILI) dataset~\cite{wuTimesNetTemporal2DVariation2023} contains records of patients experiencing severe influenza with complications.
    \item The Electricity Transformer Temperature (ETT; An indicator reflective of long-term electric power deployment)~\cite{zhouInformerEfficientTransformer2021} benchmark is comprised of two years of data, sourced from two counties in China. It comprises two hourly-level datasets (ETTh) and two 15-minute-level datasets (ETTm). Each entry within the ETT datasets includes six power load features and a target variable, termed “oil temperature”.
\end{enumerate}
Furthermore, the data statistics can be found in Table~\ref{tab:data_statistics}.

\begin{table}[h!]
    \centering
    \caption{Dataset statistics are from~\cite{wuTimesNetTemporal2DVariation2023}. The dimension indicates the number of time series (i.e., channels), and the dataset size is organized in (training, validation, testing).}
    \label{tab:data_statistics}
    \resizebox{0.95\textwidth}{!}{\begin{tabular}{l|c|c|c|c|c}
    \toprule
    Dataset & Output Length & Dataset Size & Granularity & Dim. & Domain \\
    \midrule
    Weather & \{96, 192, 336, 720\} & (36456, 5175, 10444) & 10 min & 21 & Weather \\
    Traffic & \{96, 192, 336, 720\} & (11849, 1661, 3413) & 1 hour & 862 & Transportation \\
    Electricity & \{96, 192, 336, 720\} & (17981, 2537, 5165) & 1 hour & 321 & Electricity \\
    ILI & \{24, 36, 48, 60\} & (549, 74, 170) & 1 week & 7 & Illness \\
    ETTm1, ETTm2 &  \{96, 192, 336, 720\} & (34129, 11425, 11425) & 15 min & 7 & Temperature \\
    ETTh1, ETTh2 & \{96, 192, 336, 720\} & (8209, 2785, 2785) & 1 hour & 7 & Temperature \\
    \bottomrule
    \end{tabular}}
\end{table}

\section{Implementation Details} \label{app_sec:implementation}
\subsection{Overall architecture of SPAT}
\begin{figure}
    \centering
    \includegraphics[width=0.65\linewidth]{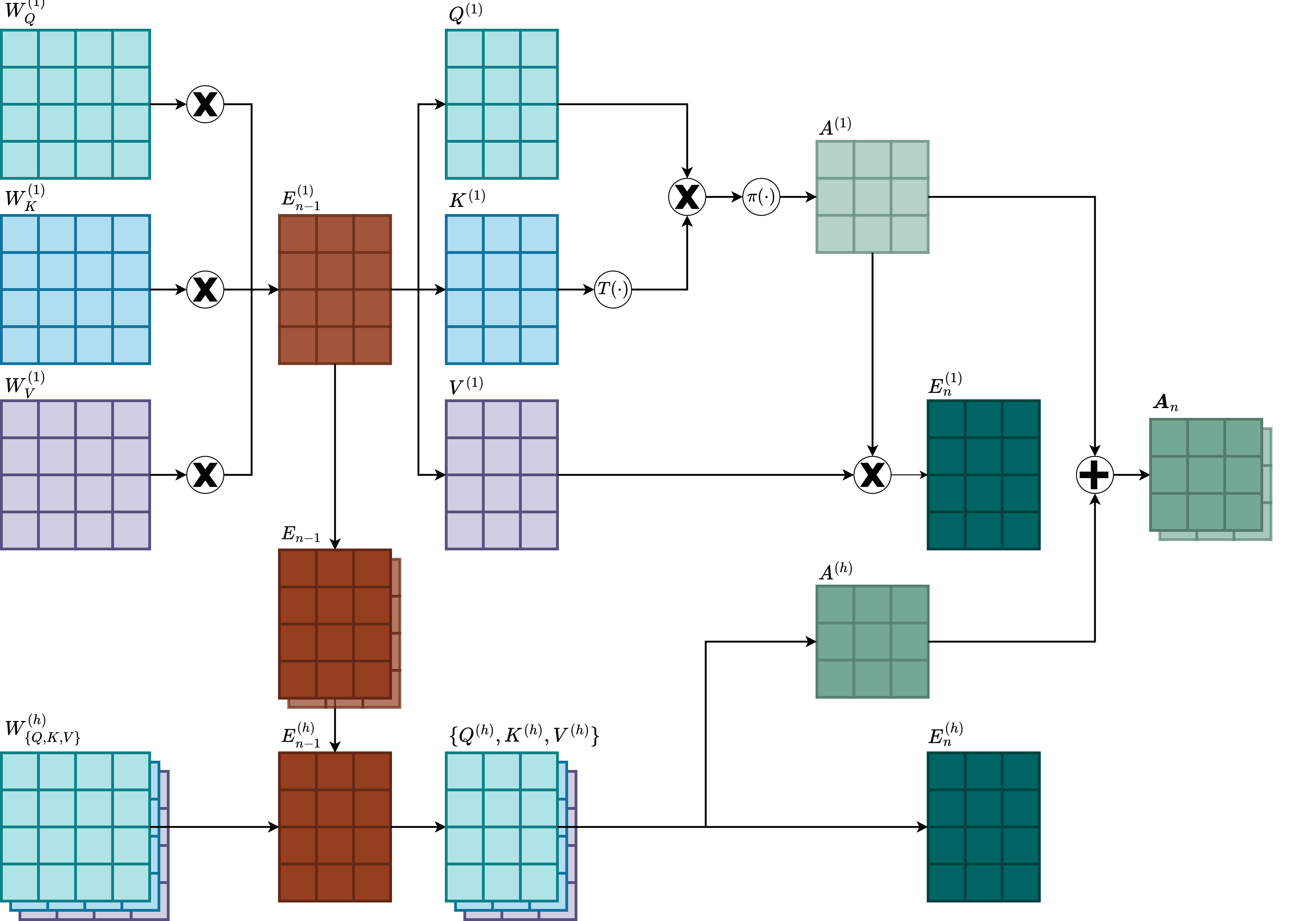}
    \caption{The MultiHead attention mechanism structure. $T(\cdot)$ and $\pi(\cdot)$ denotes the transpose and softmax operations. The upper part describes the detailed operations in each attention head, while the lower part only marks the input and output of the current head.}
    \label{fig:attention}
\end{figure}

The multihead attention module is presented in Figure \ref{fig:attention}. $E_{n-1}$ and $E_n$ are the input and output features of the attention mechanism. To measure the importance of each attention module, we utilize the attention matrix and its corresponding mask to calculate the $\bm{\mathrm{SEND}}$. 

\subsection{Details of SEND Computation}
After aggregating sensitivity matrix across different attention heads to obtain $\overline{\bm{\mathrm{Sen}}}_n \in \mathbb{R}^{L \times L}$, we compute their standard deviation of each row $i$ to get $\sigma_{n, i}$ and average these standard deviations to get $\bm{\mathrm{SEND}_n}$, shown as:
\begin{align}
    \overline{\bm{\mathrm{Sen}}}_n[i,j] &= \frac{1}{H} \sum_{h=1}^H \bm{\mathrm{Sen}}_n^{\mathrm{(norm)}}[h,i,j], \quad i,j \in \{1,...,L\},\\
    \mu_{n, i} &= \frac{1}{L}\sum_{j=1}^L \overline{\bm{\mathrm{Sen}}}_n[i,j], \\
    \sigma_{n, i} &= \sqrt{\frac{1}{L} \sum_{j=1}^L \left( \overline{\bm{\mathrm{Sen}}}_n[i,j] - \mu_{n,i} \right)^2}, \\
    \bm{\mathrm{SEND}}_n &= \frac{1}{L}\sum_{i=1}^L \sigma_{n, i}.
\end{align}

\subsection{Experiment details}
All the experiments are conducted on a single Tesla V100 GPU with 32G VRAM. The mean squared error (MSE) loss function is utilized for model optimization.
Performance comparison among different methods is conducted based on two primary evaluation metrics: Mean Squared Error (MSE) and Mean Absolute Error (MAE). 
\textbf{Mean Squared Error (MSE)}:
\begin{equation}
    \text{MSE} = \frac{1}{T} \sum_{i=1}^{T}(\mathbf{Y}_i - \hat{\mathbf{Y}}_i)^2,
\end{equation}
\textbf{Mean Absolute Error (MAE)}:
\begin{equation}
    \text{MAE} = \frac{1}{T} \sum_{i=1}^{T}|\mathbf{Y}_i - \hat{\mathbf{Y}}_i|,
\end{equation}
where \[\mathbf{Y}, \hat{\mathbf{Y}} \in \mathbb{R}^{T \times C}\] are the ground truth and prediction results of the future with $T$ time points and $C$ channels. $\mathbf{Y}_i$ denotes the $i$-th future time point.

We use the ADAM optimizer with an initial learning rate (LR) of $3 \times 10^{-4}$. This rate is modulated by a cosine learning rate scheduler. We set the $d_{\text{model}}$, $d_{\text{ff}}$ and number of MultiHead modules $N$ in alignment with the published code, and they can be found in Table~\ref{tab:appendix_configs}.

\begin{table}[t]
\centering
\caption{Training configurations for PatchTST and iTransformer.	}
\label{tab:appendix_configs}
\begin{tabular}{@{}lllll|llll@{}}
\toprule
 & \multicolumn{4}{c|}{PatchTST} & \multicolumn{4}{c}{iTransformer} \\ \cmidrule(l){2-9} 
 & LR & $d_{\text{model}}$ & $d_{\text{ff}}$ & $N$ & LR & $d_{\text{model}}$ & $d_{\text{ff}}$ & $N$ \\ \midrule
Weather & 0.0001 & 128 & 256 & 3 & 0.0001 & 512 & 512 & 3 \\
Traffic & 0.0001 & 128 & 256 & 3 & 0.001 & 512 & 512 & 4 \\
Electricity & 0.0001 & 128 & 256 & 3 & 0.0005 & 512 & 512 & 3 \\
ILI & 0.0025 & 16 & 128 & 3 & 0.0001 & 256 & 256 & 2 \\
ETTh1 & 0.0001 & 16 & 128 & 3 & 0.0001 & 256 & 256 & 2 \\
ETTh2 & 0.0001 & 16 & 128 & 3 & 0.0001 & 256 & 256 & 2 \\
ETTm1 & 0.0001 & 128 & 256 & 3 & 0.0001 & 128 & 128 & 2 \\
ETTm2 & 0.0001 & 128 & 256 & 3 & 0.0001 & 128 & 128 & 2 \\ \bottomrule
\end{tabular}
\end{table}

\section{Full Results} \label{app_sec:full_results}
In this section, we present Tables~\ref{tab:PatchTST_full} and \ref{tab:itrans_full} for PatchTST and iTransformer models, respectively, to show the full results on forecasting performances. The lookback window length is set to 336 for all datasets. We set the prediction horizon to {24, 36, 48, 60} for ILI and {96, 192, 336, 720} for others. Since we have run the experiments five times using different random seeds, we report the average results over five runs. 

\begin{table}[]
\centering
\caption{Performance of PatchTST on multivariate datasets. Results are averaged over 5 runs. K, M and G denotes Kilobytes (KB), Megabytes (MB), and Gigabytes (GB) respectively. The average result over 4 forecasting horizons is highlighted in bold.}
\label{tab:PatchTST_full}
\resizebox{\textwidth}{!}{\begin{tabular}{@{}llllllllll@{}}
\toprule
\multicolumn{2}{l}{Model} & \multicolumn{8}{c}{PatchTST} \\ \midrule
\multirow{2}{*}{} & \multirow{2}{*}{} & \multicolumn{4}{c}{Origin} & \multicolumn{4}{c}{Pruned Model} \\ 
\cmidrule(lr){3-6}  \cmidrule(lr){7-10} 
 &  & \multicolumn{1}{c}{MSE} & \multicolumn{1}{c}{MAE} & \multicolumn{1}{c}{FLOPs} & \multicolumn{1}{c}{Params} & \multicolumn{1}{c}{MSE} & \multicolumn{1}{c}{MAE} & \multicolumn{1}{c}{FLOPs} & \multicolumn{1}{c}{Params} \\ \midrule
\multirow{5}{*}{Traffic} 
 & 96 & 0.358 & 0.245 & 32.002G & 0.921M & 0.353 & 0.241 & 26.454G & 0.855M \\
 & 192 & 0.377 & 0.254 & 32.892G & 1.437M & 0.377 & 0.254 & 27.344G & 1.371M \\
 & 336 & 0.392 & 0.263 & 34.226G & 2.212M & 0.393 & 0.262 & 28.678G & 2.146M \\
 & 720 & 0.430 & 0.285 & 37.785G & 4.276M & 0.432 & 0.285 & 32.237G & 4.210M \\
 & \textbf{Avg} & \textbf{0.389} & \textbf{0.262} & \textbf{34.226G} & \textbf{2.212M} & \textbf{0.389} & \textbf{0.260} & \textbf{28.678G} & \textbf{2.146M} \\\midrule
\multirow{5}{*}{Electricity} 
 & 96 & 0.131 & 0.224 & 11.917G & 0.921M & 0.131 & 0.224 & 9.851G & 0.855M \\
 & 192 & 0.149 & 0.242 & 12.249G & 1.437M & 0.150 & 0.241 & 10.183G & 1.371M \\
 & 336 & 0.167 & 0.260 & 12.746G & 2.212M & 0.166 & 0.259 & 10.680G & 2.146M \\
 & 720 & 0.204 & 0.294 & 14.071G & 4.276M & 0.201 & 0.291 & 12.005G & 4.210M \\
 & \textbf{Avg} & \textbf{0.163} & \textbf{0.255} & \textbf{12.746G} & \textbf{2.212M} & \textbf{0.162} & \textbf{0.254} & \textbf{10.680G} & \textbf{2.146M} \\ \midrule
\multirow{5}{*}{Weather} 
 & 96 & 0.151 & 0.199 & 779.632M & 0.921M & 0.151 & 0.198 & 644.467M & 0.855M \\
 & 192 & 0.196 & 0.242 & 801.308M & 1.437M & 0.197 & 0.242 & 666.143M & 1.371M \\
 & 336 & 0.248 & 0.282 & 833.822M & 2.212M & 0.248 & 0.282 & 698.657M & 2.146M \\
 & 720 & 0.320 & 0.335 & 920.526M & 4.276M & 0.321 & 0.335 & 785.361M & 4.210M \\
 & \textbf{Avg} & \textbf{0.229} & \textbf{0.264} & \textbf{833.822M} & \textbf{2.212M} & \textbf{0.229} & \textbf{0.264} & \textbf{698.657M} & \textbf{2.146M} \\ \midrule
\multirow{5}{*}{ETTm1} 
 & 96 & 0.289 & 0.342 & 259.877M & 0.921M & 0.289 & 0.341 & 214.822M & 0.855M \\
 & 192 & 0.335 & 0.371 & 267.103M & 1.437M & 0.333 & 0.369 & 222.048M & 1.371M \\
 & 336 & 0.368 & 0.392 & 277.941M & 2.212M & 0.364 & 0.391 & 232.886M & 2.146M \\
 & 720 & 0.416 & 0.422 & 306.842M & 4.276M & 0.413 & 0.421 & 261.787M & 4.210M \\
 & \textbf{Avg} & \textbf{0.352} & \textbf{0.382} & \textbf{277.940M} & \textbf{2.212M} & \textbf{0.350} & \textbf{0.380} & \textbf{232.886M} & \textbf{2.146M} \\ \midrule
\multirow{5}{*}{ETTm2} 
 & 96 & 0.165 & 0.253 & 259.877M & 0.921M & 0.164 & 0.253 & 214.822M & 0.855M \\
 & 192 & 0.221 & 0.293 & 267.103M & 1.437M & 0.221 & 0.293 & 222.048M & 1.371M \\
 & 336 & 0.275 & 0.328 & 277.941M & 2.212M & 0.275 & 0.328 & 232.886M & 2.146M \\
 & 720 & 0.364 & 0.383 & 306.842M & 4.276M & 0.364 & 0.383 & 261.787M & 4.210M \\
 & \textbf{Avg} & \textbf{0.256} & \textbf{0.314} & \textbf{277.940M} & \textbf{2.212M} & \textbf{0.256} & \textbf{0.314} & \textbf{232.886M} & \textbf{2.146M} \\ \midrule
\multirow{5}{*}{ETTh1} 
 & 96 & 0.376 & 0.401 & 12.774M & 81.728K & 0.370 & 0.396 & 11.332M & 80.640K \\
 & 192 & 0.413 & 0.421 & 13.677M & 146.336K & 0.409 & 0.418 & 12.235M & 145.248K \\
 & 336 & 0.427 & 0.433 & 15.032M & 243.248K & 0.428 & 0.433 & 13.590M & 242.160K \\
 & 720 & 0.445 & 0.463 & 18.644M & 501.680K & 0.444 & 0.462 & 17.203M & 500.592K \\
 & \textbf{Avg} & \textbf{0.415} & \textbf{0.430} & \textbf{15.032M} & \textbf{243.248K} & \textbf{0.413} & \textbf{0.427} & \textbf{13.590M} & \textbf{242.160K} \\ \midrule
\multirow{5}{*}{ETTh2} 
 & 96 & 0.276 & 0.337 & 12.774M & 81.728K & 0.276 & 0.337 & 11.332M & 80.640K \\
 & 192 & 0.340 & 0.379 & 13.677M & 146.336K & 0.338 & 0.378 & 12.235M & 145.248K \\
 & 336 & 0.328 & 0.383 & 15.032M & 243.248K & 0.329 & 0.382 & 13.590M & 242.160K \\
 & 720 & 0.381 & 0.422 & 18.644M & 501.680K & 0.380 & 0.422 & 17.203M & 500.592K \\
 & \textbf{Avg} & \textbf{0.331} & \textbf{0.381} & \textbf{15.032M} & \textbf{243.248K} & \textbf{0.330} & \textbf{0.380} & \textbf{13.590M} & \textbf{242.160K} \\ \midrule
\multirow{5}{*}{ILI} 
 & 24 & 1.662 & 0.854 & 12.172M & 33.400K & 1.561 & 0.825 & 10.730M & 32.312K \\
 & 36 & 1.589 & 0.857 & 12.285M & 41.476K & 1.504 & 0.833 & 10.843M & 40.388K \\
 & 48 & 1.921 & 0.932 & 12.398M & 49.552K & 1.692 & 0.869 & 10.956M & 48.464K \\
 & 60 & 1.741 & 0.899 & 12.510M & 57.628K & 1.592 & 0.872 & 11.069M & 56.540K \\
 & \textbf{Avg} & \textbf{1.728} & \textbf{0.885} & \textbf{12.341M} & \textbf{45.514K} & \textbf{1.587} & \textbf{0.850} & \textbf{10.899M} & \textbf{44.426K} \\ \bottomrule
\end{tabular}}
\end{table}

\begin{table}[]
\centering
\caption{Performance of iTransformer on multivariate datasets. Results are averaged over 5 runs. K, M and G denotes Kilobytes (KB), Megabytes (MB), and Gigabytes (GB) respectively. The average result over 4 forecasting horizons is highlighted in bold.}
\label{tab:itrans_full}
\resizebox{\textwidth}{!}{\begin{tabular}{@{}llllllllll@{}}
\toprule
\multicolumn{2}{l}{Model} & \multicolumn{8}{c}{PatchTST} \\ \midrule
\multirow{2}{*}{} & \multirow{2}{*}{} & \multicolumn{4}{c}{Origin} & \multicolumn{4}{c}{Pruned Model} \\ 
\cmidrule(lr){3-6}  \cmidrule(lr){7-10} 
 &  & \multicolumn{1}{c}{MSE} & \multicolumn{1}{c}{MAE} & \multicolumn{1}{c}{FLOPs} & \multicolumn{1}{c}{Params} & \multicolumn{1}{c}{MSE} & \multicolumn{1}{c}{MAE} & \multicolumn{1}{c}{FLOPs} & \multicolumn{1}{c}{Params} \\ \midrule
\multirow{5}{*}{Traffic} 
 & 96 & 0.356 & 0.258 & 17.447G & 6.535M & 0.352 & 0.253 & 14.095G & 5.484M \\
 & 192 & 0.376 & 0.268 & 17.533G & 6.584M & 0.365 & 0.262 & 14.180G & 5.533M \\
 & 336 & 0.389 & 0.274 & 17.660G & 6.658M & 0.376 & 0.267 & 14.308G & 5.607M \\
 & 720 & 0.423 & 0.290 & 18.001G & 6.855M & 0.405 & 0.280 & 14.649G & 5.804M \\
 & \textbf{Avg} & \textbf{0.386} & \textbf{0.273} & \textbf{17.660G} & \textbf{6.658M} & \textbf{0.375} & \textbf{0.266} & \textbf{14.308G} & \textbf{5.607M} \\ \midrule
\multirow{5}{*}{Electricity} 
 & 96 & 0.131 & 0.227 & 3.867G & 4.957M & 0.129 & 0.222 & 1.173G & 1.804M \\
 & 192 & 0.155 & 0.250 & 3.899G & 5.006M & 0.148 & 0.239 & 1.209G & 1.854M \\
 & 336 & 0.170 & 0.267 & 3.947G & 5.080M & 0.165 & 0.257 & 1.253G & 1.928M \\
 & 720 & 0.202 & 0.294 & 4.075G & 5.277M & 0.202 & 0.290 & 1.381G & 2.125M \\
 & \textbf{Avg} & \textbf{0.165} & \textbf{0.259} & \textbf{3.947G} & \textbf{5.080M} & \textbf{0.161} & \textbf{0.252} & \textbf{1.253G} & \textbf{1.928M} \\ \midrule
\multirow{5}{*}{Weather} 
 & 96 & 0.160 & 0.209 & 251.353M & 4.957M & 0.151 & 0.199 & 90.227M & 1.805M \\
 & 192 & 0.206 & 0.251 & 253.811M & 5.006M & 0.193 & 0.241 & 92.685M & 1.854M \\
 & 336 & 0.257 & 0.290 & 257.498M & 5.080M & 0.245 & 0.282 & 96.371M & 1.928M \\
 & 720 & 0.329 & 0.340 & 267.328M & 5.277M & 0.323 & 0.336 & 106.202M & 2.125M \\
 & \textbf{Avg} & \textbf{0.238} & \textbf{0.273} & \textbf{257.498M} & \textbf{5.080M} & \textbf{0.228} & \textbf{0.265} & \textbf{96.371M} & \textbf{1.928M} \\ \midrule
\multirow{5}{*}{ETTm1} 
 & 96 & 0.304 & 0.357 & 5.707M & 254.944K & 0.292 & 0.346 & 2.699M & 122.848K \\
 & 192 & 0.342 & 0.380 & 5.977M & 267.328K & 0.331 & 0.369 & 2.970M & 135.232K \\
 & 336 & 0.382 & 0.403 & 6.383M & 285.904K & 0.367 & 0.390 & 3.375M & 153.808K \\
 & 720 & 0.443 & 0.439 & 7.464M & 335.440K & 0.425 & 0.423 & 4.456M & 203.344K \\
 & \textbf{Avg} & \textbf{0.368} & \textbf{0.395} & \textbf{6.382M} & \textbf{285.904K} & \textbf{0.354} & \textbf{0.382} & \textbf{3.375M} & \textbf{153.808K} \\ \midrule
\multirow{5}{*}{ETTm2} 
 & 96 & 0.178 & 0.268 & 5.707M & 254.944K & 0.174 & 0.264 & 2.699M & 122.848K \\
 & 192 & 0.244 & 0.313 & 5.977M & 267.328K & 0.236 & 0.306 & 2.970M & 135.232K \\
 & 336 & 0.297 & 0.347 & 6.383M & 285.904K & 0.290 & 0.340 & 3.375M & 153.808K \\
 & 720 & 0.383 & 0.398 & 7.463M & 335.440K & 0.369 & 0.390 & 4.456M & 203.344K \\
 & \textbf{Avg} & \textbf{0.275} & \textbf{0.331} & \textbf{6.382M} & \textbf{285.904K} & \textbf{0.267} & \textbf{0.325} & \textbf{3.375M} & \textbf{153.808K} \\ \midrule
\multirow{5}{*}{ETTh1} 
 & 96 & 0.405 & 0.419 & 20.064M & 0.903M & 0.383 & 0.405 & 8.282M & 0.377M \\
 & 192 & 0.448 & 0.447 & 20.605M & 0.928M & 0.419 & 0.428 & 8.823M & 0.401M \\
 & 336 & 0.475 & 0.466 & 77.434M & 3.502M & 0.438 & 0.440 & 30.801M & 1.401M \\
 & 720 & 0.557 & 0.534 & 81.760M & 3.699M & 0.462 & 0.477 & 35.127M & 1.598M \\
 & \textbf{Avg} & \textbf{0.471} & \textbf{0.466} & \textbf{49.966M} & \textbf{2.258M} & \textbf{0.426} & \textbf{0.438} & \textbf{20.758M} & \textbf{0.944M} \\ \midrule
\multirow{5}{*}{ETTh2} 
 & 96 & 0.306 & 0.361 & 5.707M & 254.944K & 0.292 & 0.351 & 2.699M & 122.848K \\
 & 192 & 0.380 & 0.407 & 5.977M & 267.328K & 0.357 & 0.391 & 2.970M & 135.232K \\
 & 336 & 0.425 & 0.437 & 6.383M & 285.904K & 0.383 & 0.414 & 3.375M & 153.808K \\
 & 720 & 0.427 & 0.449 & 7.464M & 335.440K & 0.412 & 0.442 & 4.456M & 203.344K \\
 & \textbf{Avg} & \textbf{0.384} & \textbf{0.414} & \textbf{6.382M} & \textbf{285.904K} & \textbf{0.361} & \textbf{0.399} & \textbf{3.375M} & \textbf{153.808K} \\ \midrule
\multirow{5}{*}{ILI} 
 & 24 & 2.307 & 1.037 & 18.352M & 0.825M & 2.175 & 0.988 & 6.580M & 0.298M \\
 & 36 & 2.238 & 1.023 & 18.419M & 0.828M & 2.175 & 1.005 & 6.637M & 0.302M \\
 & 48 & 2.126 & 1.001 & 71.577M & 3.235M & 2.032 & 0.968 & 24.944M & 1.134M \\
 & 60 & 2.138 & 1.024 & 71.712M & 3.242M & 2.062 & 0.993 & 25.079M & 1.140M \\
 & \textbf{Avg} & \textbf{2.202} & \textbf{1.021} & \textbf{45.015M} & \textbf{2.0320M} & \textbf{2.111} & \textbf{0.989} & \textbf{15.808M} & \textbf{0.719M} \\ \bottomrule
\end{tabular}}
\end{table}

\section{Limitations and Future Works} \label{app_sec:limitation}
While the SPAT-pruned models demonstrate significant improvements in multivariate time series forecasting, several limitations must be acknowledged, providing directions for future work.

\subsection{Structured Pruning Based on Finer-Grained Units}
Our work treats the multi-head attention (MHA) mechanism as the pruning unit, encompassing the linear projections of $Q$, $K$, and $V$, the computation of attention scores, and the final value projection. However, each MHA module consists of multiple heads, and different heads may capture diverse types of information. This suggests that pruning at the attention-head level, which is a finer granularity, may offer further improvements in forecasting performance by more precisely removing redundant or uninformative components.

\subsection{Limited Task Scope}
Our proposed pruning algorithm has been thoroughly evaluated only within the context of time series forecasting. While attention-based models are widely used in other domains, such as image classification in computer vision (CV) and question answering in natural language processing (NLP), these tasks remain unexplored in our current study. As attention mechanisms are fundamental to models like ViT~\cite{dosovitskiyImageWorth16x162021a} and LLaMA-7B~\cite{touvronLlama2Open2023}, future work will aim to extend the SPAT algorithm to these domains, ensuring that only the most relevant attention modules are preserved during inference across various modalities.

\section{Societal Impacts} \label{app_sec:societal_impacts}
The development of the SPAT (Sensitivity Pruner for Attention) algorithm holds significant potential for improving time series forecasting efficiency across various domains, including healthcare, transportation, finance, and energy management. By enabling the reduction of computational overhead while maintaining or improving forecasting performance, SPAT can help organizations make faster, more cost-effective decisions. This could lead to better resource allocation, improved planning, and more responsive systems, ultimately benefiting industries that rely heavily on predictive analytics. However, there are also potential societal challenges to consider. Excessive reliance on automated pruning techniques might lead to over-simplified models that overlook important nuances in certain datasets, potentially introducing biases or inaccurate forecasts. To mitigate this, it is essential to ensure that SPAT is applied responsibly, with regular evaluation and oversight to validate its accuracy and fairness across different contexts. Additionally, maintaining transparency in the model’s decision-making process will be crucial to foster trust and ensure that its benefits are realized across diverse sectors without unintended consequences.

\end{document}